\pdfoutput=1

\documentclass[11pt]{article}

\usepackage[]{ACL2023}

\usepackage{times}
\usepackage{latexsym}
\usepackage[T1]{fontenc}
\usepackage{ulem}
\usepackage{xcolor}
\usepackage[utf8]{inputenc}

\usepackage{microtype}
\usepackage{booktabs}

\usepackage{inconsolata}

\usepackage{graphicx}

\usepackage{subfig}

\title{A Technical Report for Polyglot-Ko: Open-Source Large-Scale Korean Language Models}

    \author{Hyunwoong Ko\thanks{\;\;Equal contribution}\;\;, Kichang Yang\footnotemark[1]\;\;, Minho Ryu\footnotemark[1]\;\;, Taekyoon Choi\footnotemark[1] \\  \textbf{Seungmu Yang\footnotemark[1]\;\;}, \textbf{Jiwung Hyun\footnotemark[1]\;\;, Sungho Park\footnotemark[1]\;\;, Kyubyong Park\thanks{\;\;Affiliation: TUNiB}} \\\\EleutherAI}

\begin{document}
\maketitle
\begin{abstract}
Polyglot is a pioneering project aimed at enhancing the non-English language performance of multilingual language models. Despite the availability of various multilingual models such as mBERT \cite{devlin2019bert}, XGLM \cite{lin2022fewshot}, and BLOOM \cite{scao2022bloom}, researchers and developers often resort to building monolingual models in their respective languages due to the dissatisfaction with the current multilingual models' non-English language capabilities. Addressing this gap, we seek to develop advanced multilingual language models that offer improved performance in non-English languages.
In this paper, we introduce the Polyglot Korean models, which represent a specific focus rather than being multilingual in nature. In collaboration with TUNiB\footnote{\url{https://tunib.ai/}}, our team collected 1.2TB of Korean data meticulously curated for our research journey. We made a deliberate decision to prioritize the development of Korean models before venturing into multilingual models. This choice was motivated by multiple factors: firstly, the Korean models facilitated performance comparisons with existing multilingual models; and finally, they catered to the specific needs of Korean companies and researchers.
This paper presents our work in developing the Polyglot Korean models, which propose some steps towards addressing the non-English language performance gap in multilingual language models.
\phantom{Polyglot is a project aimed at developing multilingual language models with higher non-English language performance. Despite the availability of various multilingual models such as mBERT, BLOOM, and XGLM, people around the world still develop monolingual models in their respective languages. One of the significant reasons behind this trend is the dissatisfaction with the non-English language performance of current multilingual models. Hence, there is a need to develop new multilingual models with improved non-English language performance. In this regard, We seeks to bridge the existing gap and develop advanced multilingual language models. However, in this paper, we introduce Polyglot Korean models, which are not multilingual. We started our research with 1.2TB of Korean data collected by TUNiB and decided to develop Korean models before multilingual models. This decision was based on the fact that Korean models would allow us to practice multi-node training, enable performance comparison with multilingual models, and be useful to Korean companies and researchers. Therefore, in this paper, we present our work on developing Polyglot Korean models.}
\end{abstract}

\section{Introduction}

The advent of large-scale language models has revolutionized the field of natural language processing, leading to significant advancements in various applications, such as language translation and text classification \cite{devlin2019bert,radford2019language,liu2019roberta,clark2020electra,chowdhery2022palm,anil2023palm}. While numerous large language models for English have been publicly released \cite{zhang2022opt,black2022gptneox20b,biderman2023pythia,touvron2023llama,together2023redpajama,MosaicML2023Introducing}, the availability of such models for non-English languages remains limited. Although several multilingual large language models have also been released \cite{lin2022fewshot,scao2022bloom}, they are typically trained on English-centric corpora, resulting in lower performance on other languages. 

However, with the increasing interest in non-English languages, there is a growing need for high-performance language models that are specifically tailored to these languages. To tackle this challenge, we initiated the Polyglot project, which focuses on the development of large language models customized for non-English languages. As part of the project, our first model in this endeavor is Polyglot-Ko, an exceptional language model specifically designed for the Korean language. We chose to prioritize the Korean language as the initial model because our founding members are primarily Korean and we had a readily available dataset for training purposes. Our goal is to make these language models accessible to researchers and practitioners, empowering them to explore and advance natural language processing tasks in their respective languages. Polyglot-Ko leverages the transformer architecture, known for its effectiveness in capturing long-range dependencies in natural language text. Our model has been trained on an extensive corpus of text data, incorporating diverse sources such as web pages, news articles, and social media posts to ensure its versatility across different domains and styles.

In this technical report, we provide a comprehensive description of the architecture and training process for four distinct Polyglot-Ko models, differing in parameter sizes: 1.3B (billion), 3.8B, 5.8B, and 12.8B. Notably, our 12.8 billion parameter model represents the largest publicly available Korean language model suitable for commercial applications, making it an invaluable resource for researchers and practitioners engaged in Korean natural language processing tasks.

We assess the zero-shot and few-shot performance of our Polyglot-Ko models using the KOBEST benchmark \cite{kim2022kobest}. Through our experiments, we have successfully demonstrated that Polyglot-Ko attains competitive results across various benchmark datasets.

In addition to presenting our achievements, we also acknowledge potential limitations and identify areas that warrant future improvement. By offering recommendations for further research, we aspire to foster advancements in the field. We firmly believe that Polyglot-Ko will serve as a valuable resource for the Korean natural language processing community, enabling the development of innovative applications and contributing to a deeper understanding of the intricacies and dynamics of the Korean language.

\section{Datasets}
We collaborated with TUNiB to collect a large-scale Korean language dataset for our research. The dataset, totaling 1.2TB, was meticulously gathered through our collaborative efforts. Subsequently, we performed preprocessing on this dataset, resulting in 863GB of text data that served as the foundation for our analysis and model training.

\begin{table}[htbp]
\centering
\begin{tabular}{lc}
\hline
\textbf{Source} & \textbf{Size (GB)} \\
\hline
Korean blog posts & 682.3 \\
Korean news dataset & 87.0 \\
\href{https://corpus.korean.go.kr}{Modu corpus} & 26.4 \\
Korean patent dataset & 19.0 \\
Korean Q\&A dataset & 18.1 \\
\href{https://github.com/Beomi/KcBERT}{KcBert dataset} & 12.7 \\
Korean fiction dataset & 6.1\\
Korean online comments & 4.2\\
\href{https://ko.wikipedia.org}{Korean wikipedia} & 1.4 \\
\href{https://github.com/clovaai/ClovaCall}{Clova call} & $<$ 1.0 \\
\href{https://github.com/e9t/nsmc}{Naver Sentiment Movie Corpus} & $<$ 1.0 \\
Korean hate speech dataset & $<$ 1.0 \\
\href{https://opus.nlpl.eu/OpenSubtitles.php}{Open subtitles} & $<$ 1.0 \\
\href{https://aihub.or.kr}{AIHub various tasks datasets} & $<$ 1.0 \\
\href{https://stdict.korean.go.kr/main/main.do}{Standard Korean dictionary} & $<$ 1.0 \\
\hline
\end{tabular}
\caption{Datasets for the Korean language.}
\label{table:korean_datasets}
\end{table}


\subsection{Data Analysis}
The primary motivation behind our data analysis is to mitigate potential risks that may arise during both training and inference stages. We strive to overcome various issues that can adversely impact model performance and reliability. For instance, we identify and handle challenges such as empty or excessively short text data, repeated words and characters, and instances of duplicated data, which can pose problems during model training. Additionally, we pay particular attention to the inference stage, where the presence of personally identifiable information (PII) can create complications. By thoroughly analyzing the data, we strive to minimize these risks and ensure the robustness and privacy compliance of our models.
Through a meticulous examination of the data with the aim of mitigating these risks, we successfully categorized them into four distinct types:
\begin{itemize}
\item \textbf{Data available for training}: This category predominantly comprises news and Wikipedia data, which provide substantial information with sufficiently long text sequences.
\item \textbf{Data requiring contextual information for training}: In this category, we primarily encountered blog data and news data. These datasets contained numerous short texts that were incorrectly scraped, necessitating the inclusion of contextual information during the training process.
\item \textbf{Data containing hate speech}: We observed a significant presence of hate speech in datasets sourced from certain community websites, highlighting the importance of addressing this issue during model training.
\item \textbf{NLP task-specific data}: This category encompasses data specifically designed for NLP tasks such as text classification or entity recognition. While this data can be utilized for model training, it necessitates separate handling during model evaluation.
\end{itemize}

During the examination of data types, we encountered various quality issues that required preprocessing to ensure optimal learning. Recognizing the potential detrimental impact of these issues on model training, we organized them and incorporated them into our text preprocessing pipeline. The specific quality issues we addressed include:

\begin{itemize}
\item \textbf{Empty text}: Instances with no text content.
\item \textbf{Unnecessary spaces}: Instances containing multiple unnecessary spaces.
\item \textbf{De-identification}: Identification and removal of personally identifiable information within the data instances.
\item \textbf{Uncleaned HTML tags}: Removal of HTML tags that had not been properly cleaned.
\item \textbf{Deduplication}: Identification and removal of duplicated data instances based on exact matches.
\item \textbf{Broken code}: Handling instances where only fragments of HTML or Markdown were present.
\item \textbf{Short text}: Detection and handling of excessively short data instances.
\item \textbf{Repeated characters}: Identification and addressing of instances with repeated characters.
\end{itemize}

By addressing these quality issues as part of our text preprocessing workflow, we aimed to enhance the quality and reliability of the data used for model training.
One crucial aspect of our data preprocessing was the removal of HTML-like code to a large extent, as the focus of the polyglot model is on generating Korean text. Additionally, the length of the data played a significant role in our preprocessing efforts. Longer text lengths provide more contextual information for model training, while shorter text lengths limit the available context for text generation. 

\begin{table*}
    \centering
    \begin{tabular}{lcccc}
        \toprule
            Hyperparameter & \textbf{1.3B} & \textbf{3.8B} & \textbf{5.8B} & \textbf{12.8B} \\
        \midrule
        \(n_{parameters}\) & 1,331,810,304 & 3,809,974,272 & 5,885,059,072 & 12,898,631,680 \\
        \(n_{layers}\) & 24 & 32 & 28 & 40 \\
        \(d_{model}\) & 2,048 & 3,072 & 4,096 & 5,120 \\
        \(d_{ff}\) & 8,192 & 12,288 & 16,384 & 20,480 \\
        \(n_{heads}\) & 16 & 24 & 16 & 40 \\
        \(d_{head}\) & 128 & 128 & 256 & 128 \\
        \(n_{vocab}\) & 30,003 / 30,080 & 30,003 / 30,080 & 30,003 / 30,080 & 30,003 / 30,080 \\
        Positional Encoding & \href{https://arxiv.org/abs/2104.09864}{Rotary (RoPE)} & Rotary (RoPE) & Rotary (RoPE) & Rotary (RoPE) \\
        RoPE Dimensions & \href{https://github.com/kingoflolz/mesh-transformer-jax/blob/f2aa66e0925de6593dcbb70e72399b97b4130482/mesh_transformer/layers.py#L223}{64} 
 & 64 & 64 & 64\\
        \bottomrule
    \end{tabular}
    \caption{The configuration settings of the Polyglot-Ko model.}
    \label{tab:configures}
\end{table*}

\section{Models}
To train the polyglot models, we used EleutherAI's GPT-NeoX codebase \cite{gpt-neox-library}. This codebase provided a solid foundation for our training process. Additionally, we received invaluable assistance from Stability AI\footnote{\url{https://stability.ai/}}, who provided access to 256 A100s (8 * 32 nodes) on HPC clusters\footnote{\url{https://hpc.stability.ai/}}. This computational infrastructure played a crucial role in efficiently training our models.
The training tokens information for each model is as follows:
\begin{itemize}
\item The 1.3B model was trained on 213B tokens.
\item The 3.8B model was trained on 219B tokens.
\item The 5.8B model was trained on 172B tokens.
\item The 12.8B model was trained on 167B tokens.
\end{itemize}

Due to the limitations of available computational resources, the models were trained with varying numbers of tokens. Despite our efforts \footnote{We suspect it was overtrained or overfitting, but there are not any sciencific evidence.}, moreover, broken generation (e.g generate same tokens repeatedly) occurred as the loss sharply dropped near the 1 epoch boundary for 1.3B and 3.8B models. Consequently, we made the decision to select and early stop the model checkpoints prior to the epoch boundary, as they exhibited better generation performance.

A consistent tokenizer was utilized across all models, featuring a vocabulary size of 30003. The tokenizer was trained using morpheme-aware Byte-Level BPE (Byte-Pair Encoding). We performed morpheme analysis using MeCab\footnote{\url{https://bitbucket.org/eunjeon/mecab-ko-dic/src/master/}}, a widely-used morphological analysis tool for Korean text. This ensured that the models were equipped with an effective and consistent tokenization scheme tailored to the Korean language.

Now, let's delve into the details of each model's training process:

\paragraph{1.3B Model}
The 1.3B model was trained without model parallelism, utilizing a total batch size of 1024. However, broken generation was observed as the loss sharply dropped around the 100,000 steps. To handle this, model checkpoints were evaluated and verified prior to the onset of it to ensure optimal model selection.

\paragraph{3.8B Model}
Similar to the 1.3B model, the 3.8B model also experienced same symptoms at around 100,000 steps. Model parallelism was applied during training, and the overall batch size remained the same as that of the 1.3B model. As a result, the decision was made to halt the model training process.

\paragraph{5.8B Model}
The 5.8B model utilized model parallelism, and the overall batch size was reduced by 1/4 compared to the 1.3B and 3.8B models. We trained with 172B tokens, which consisted of a total of 320,000 steps. As it was trained with lower tokens compared with 1.3B and 3.8B, The model's performance consistently improved as the number of training steps increased.

\paragraph{12.8B Model}
For the 12.8B model, model parallelism was employed, with a scale of 2 times larger than the 5.8B model. The overall batch size was maintained through the use of gradient accumulation steps (GAS). The model was trained for a total of 301,000 steps.

\section{Experiments}

\begin{table*}
    \centering
    \begin{tabular}{lccccccccc}
        \toprule
        \multicolumn{2}{c}{} & \multicolumn{4}{c}{\textbf{COPA} (n=shot)} & \multicolumn{4}{c}{\textbf{HellaSwag}} \\
        \cmidrule(lr){3-6} \cmidrule(lr){7-10}
        \textbf{Model} & \textbf{Params} & n=0 & n=5 & n=10 & n=50 & n=0 & n=5 & n=10 & n=50 \\
        \midrule
            Ko-GPT-Trinity & 1.2B & 0.670 & 0.648 & 0.642 & 0.651 & 0.524 & 0.527 & 0.517 & 0.535 \\
            Polyglot-Ko (ours) & 1.3B & 0.720 & 0.719 & 0.720 & 0.721 & 0.525 & 0.526 & 0.528 & 0.543 \\
            Polyglot-Ko (ours) & 3.8B & 0.760 & 0.761 & 0.764 & 0.779 & 0.571 & 0.583 & 0.567 & 0.579 \\
            Polyglot-Ko (ours) & 5.8B & 0.775 & 0.768 & 0.778 & 0.789 & 0.598 & 0.600 & 0.598 & \textbf{0.621} \\
            KoGPT & 6.0B & 0.735 & 0.729 & 0.728 & 0.748 & 0.559 & 0.583 & 0.583 & 0.591 \\
            xGLM & 7.5B & 0.672 & 0.673 & 0.677 & 0.712 & 0.566 & 0.569 & 0.556 & 0.562 \\
            Polyglot-Ko (ours) & 12.8B & \textbf{0.794} & \textbf{0.811} & \textbf{0.804} & \textbf{0.837} & \textbf{0.595} & \textbf{0.631} & \textbf{0.610} & 0.612 \\
        \bottomrule
    \end{tabular}
    \caption{Comparative results of the performance (F1 score) on the COPA and HellaSwag tasks.}
    \label{tab:results}
\end{table*}

\begin{table*}
    \centering
    \begin{tabular}{lccccccccc}
        \toprule
        \multicolumn{2}{c}{} & \multicolumn{4}{c}{\textbf{BoolQ} (n=shot)} & \multicolumn{4}{c}{\textbf{SentiNeg}} \\
        \cmidrule(lr){3-6} \cmidrule(lr){7-10}
        \textbf{Model} & \textbf{Params} & n=0 & n=5 & n=10 & n=50 & n=0 & n=5 & n=10 & n=50 \\
        \midrule
            Ko-GPT-Trinity & 1.2B & 0.336 & 0.401 & 0.364 & 0.356 & 0.606 & 0.688 & 0.728 & 0.841 \\
            Polyglot-Ko (ours) & 1.3B & 0.355 & 0.475 & 0.411 & 0.404 & 0.679 & 0.626 & 0.551 & 0.785 \\
            Polyglot-Ko (ours) & 3.8B & 0.432 & 0.526 & 0.493 & 0.404 & 0.486 & 0.795 & 0.732 & 0.785 \\
            Polyglot-Ko (ours) & 5.8B & 0.436 & 0.570 & 0.519 & 0.524 & 0.339 & 0.884 & 0.881 & 0.952 \\
            KoGPT & 6.0B & 0.451 & 0.598 & 0.550 & 0.520 & 0.375 & 0.894 & 0.929 & 0.970 \\
            xGLM & 7.5B & 0.446 & 0.332 & 0.332 & 0.332 & 0.358 & 0.447 & 0.396 & 0.527 \\
            Polyglot-Ko (ours) & 12.8B & \textbf{0.482} & \textbf{0.604} & \textbf{0.629} & \textbf{0.645} & \textbf{0.912} & \textbf{0.902} & \textbf{0.934} & \textbf{0.972} \\
        \bottomrule
    \end{tabular}
    \caption{Comparative results of the performance (F1 score) on the BoolQ and SentiNeg tasks.}
    \label{tab:results_bq_sn}
\end{table*}

\begin{figure*}
  \centering
  \subfloat{\includegraphics[width=.49\textwidth]{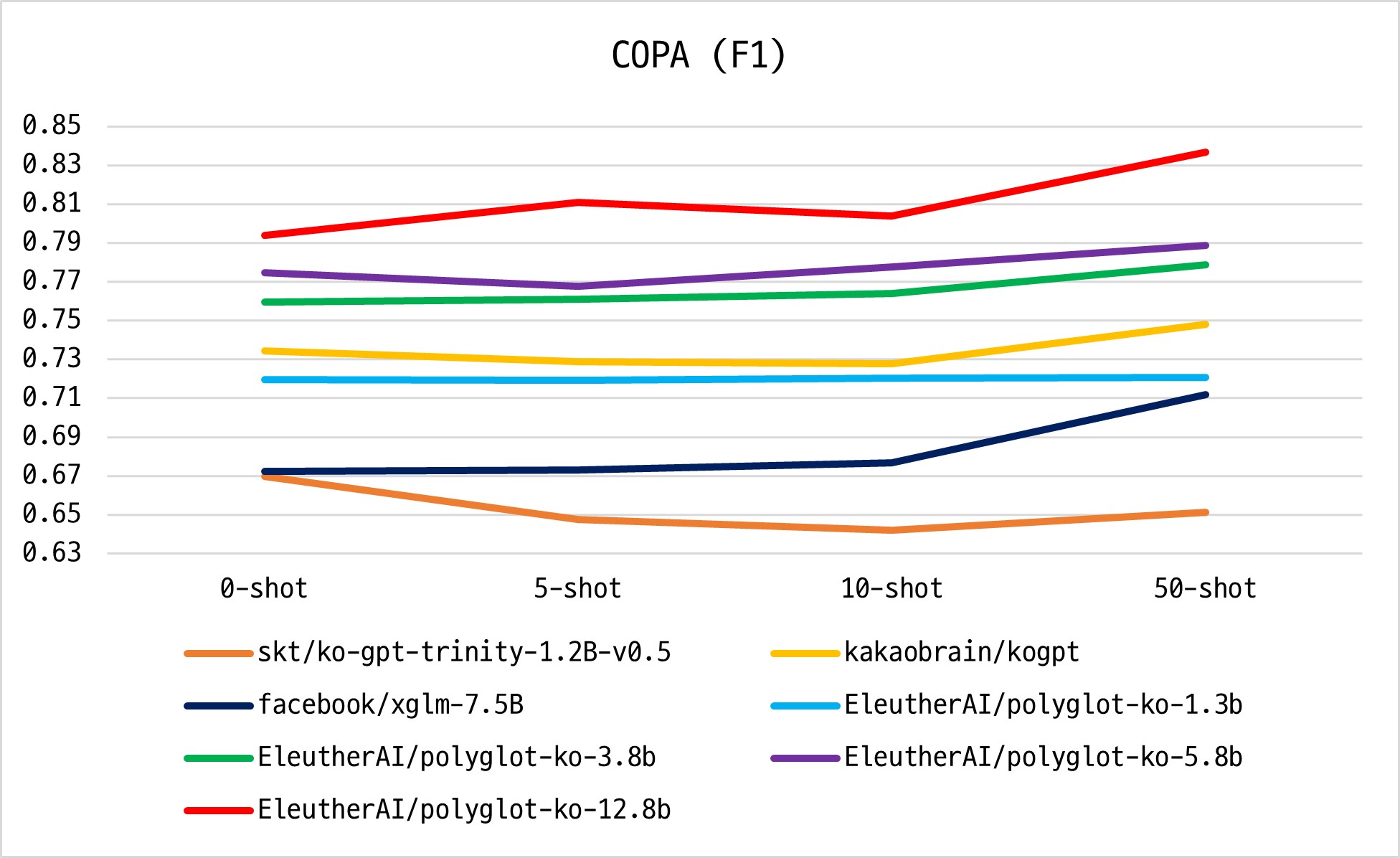}\label{fig:subfig1}}
  \hfill
  \subfloat{\includegraphics[width=.49\textwidth]{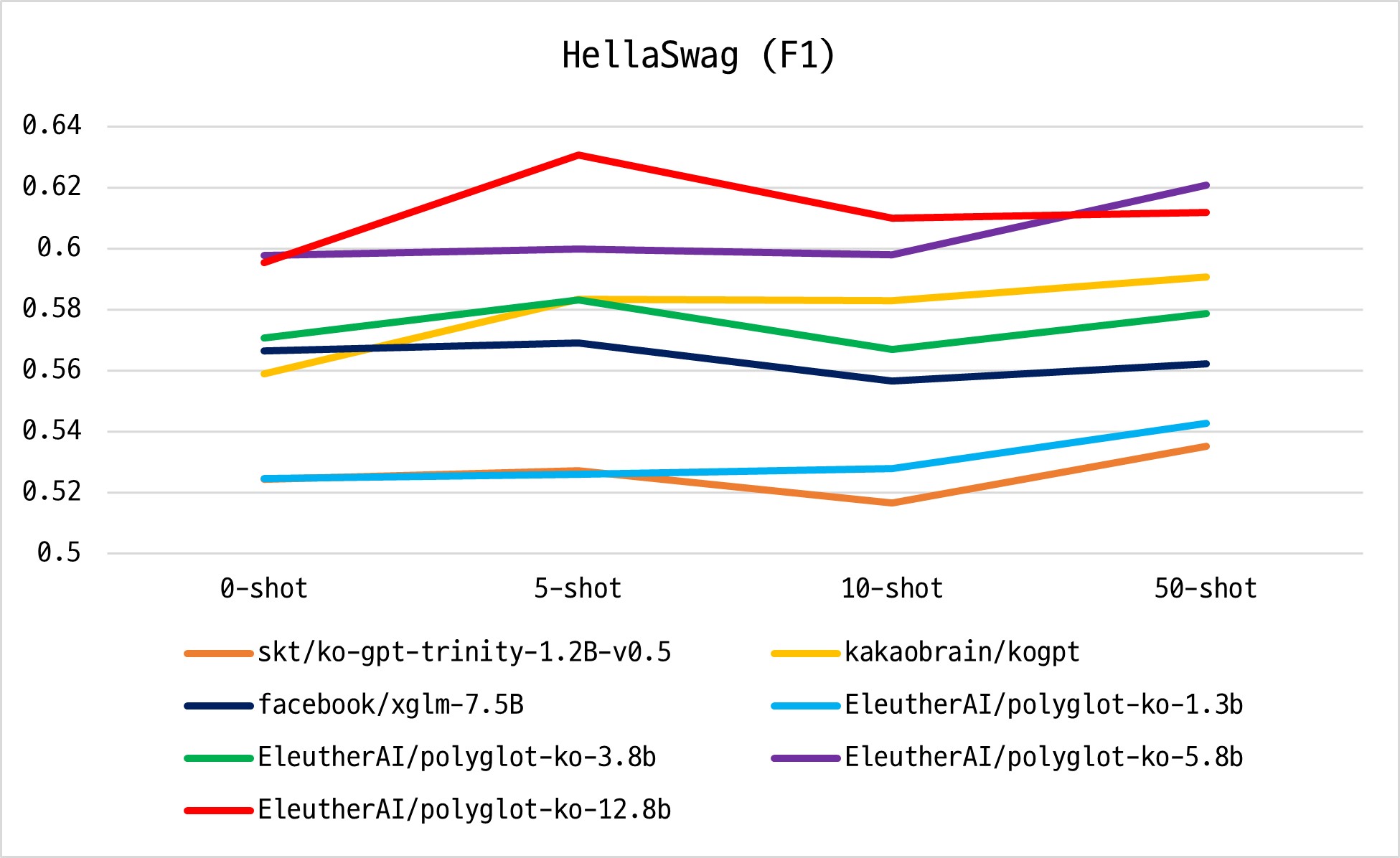}\label{fig:subfig2}} \\
  \vspace{\baselineskip}
  \subfloat{\includegraphics[width=.49\textwidth]{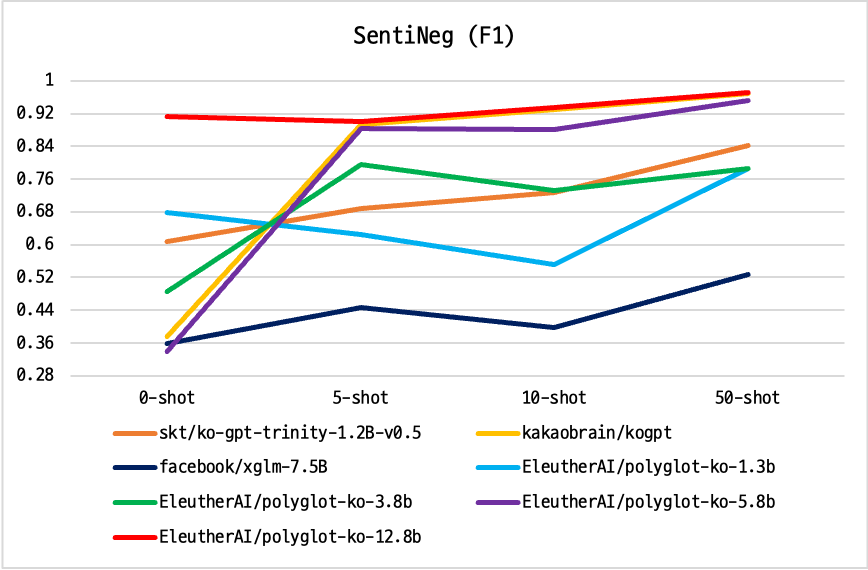}\label{fig:subfig3}}
  \hfill
  \subfloat{\includegraphics[width=.49\textwidth]{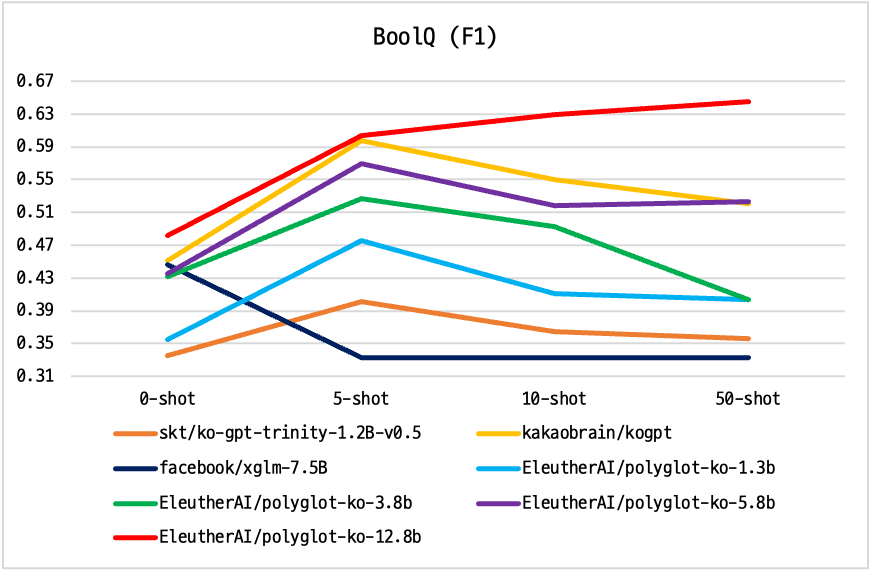}\label{fig:subfig4}}
  \caption{The performance metrics of COPA (top left), HellaSwag (top right), SentiNeg (bottom left), and BoolQ (bottom right) tasks using the KOBEST dataset, with all metrics measured using the F1 score.}
  \label{fig:whole-figure}
\end{figure*}

\begin{figure}
  \begin{minipage}[t]{.48\textwidth}
    \centering
    \includegraphics[width=\textwidth]{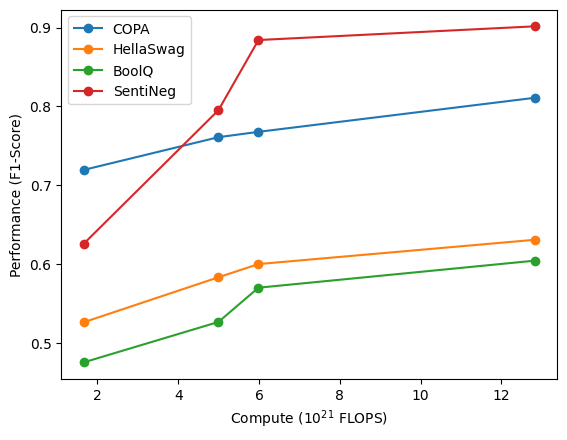}
    \caption{The 5-shot performance of Polyglot-Ko models on each task demonstrates a clear trend that as the compute increases, the performance improves.}
    \label{fig:5-shot}
  \end{minipage}
\end{figure}

We conducted an evaluation of Polyglot-Ko on the KOBEST dataset \cite{kim2022kobest}, which encompasses five downstream tasks: COPA, HellaSwag, SentiNeg, BoolQ, and WiC. Each task focuses on different aspects of language understanding and reasoning.

\begin{itemize}
\item \textbf{COPA} requires the selection of an alternative that is a cause/effect of a given premise.
\item \textbf{HellaSwag} evaluates commonsense reasoning and natural language inference.
\item \textbf{BoolQ} is designed to test the models' ability to answer questions that require reasoning over multiple sentences.
\item \textbf{SentiNeg} focuses on sentiment analysis in the Korean language.
\item \textbf{WiC} requires identifying whether the meaning of a target word is the same or different in two given contexts.
\end{itemize}

During the evaluation, we compared the performance of Polyglot-Ko with other similar models, including ko-gpt-trinity-1.2B\footnote{\url{https://huggingface.co/skt/ko-gpt-trinity-1.2B-v0.5}}, KoGPT \cite{kakaobrain2021kogpt}, and XGLM-7.5B \cite{lin2022fewshot}. These models were chosen as they are currently the only publicly available billion-scale Korean language models, excluding other multilingual models. The evaluation process involved using the provided prompts, and F1-scores were used as the evaluation metric for all tasks. We utilized the polylgot\footnote{\url{https://github.com/EleutherAI/lm-evaluation-harness/tree/polyglot}} branch of EleutherAI's lm-evaluation-harness \cite{eval-harness} repository, which provided the evaluation codebase for our experiments. This allowed for consistent and standardized evaluation procedures across different models, ensuring fair comparisons.

Note that we have chosen not to report the performance of Polyglot-Ko on the WiC task in this section due to nearly random performance observed across all models. However, detailed results for the WiC task can be found in the Appendix \ref{sec:appendix-wic}.
Figure \ref{fig:whole-figure} illustrates the results obtained when varying the number of few-shot examples.

\subsection{COPA}
The experiment was conducted on the COPA task, and the results are presented in the left portion of Table \ref{tab:results}. The table provides a comprehensive comparison of model performance based on the number of few-shot examples. To ensure fairness, all models were evaluated under the same conditions and using identical prompts.
The results clearly demonstrate that our 12.8B model outperforms the other models across all scenarios. Specifically, our 12.8B model achieves the highest F1 score in both 0-shot and 50-shot settings. For 0-shot, our model achieves an impressive F1 score of 0.7937, surpassing all other models. Similarly, in the 50-shot scenario, our 12.8B model achieves the highest F1 score of 0.8368.
These results highlight the superiority of our 12.8B model over comparable models in the COPA task. It showcases the model's exceptional performance and reinforces its effectiveness in understanding and reasoning tasks.

\subsection{HellaSwag}
The right portion of Table \ref{tab:results} presents the performance of different language models on the HellaSwag task for the Korean language. The table includes the model name, the number of parameters in each model, and their respective performance at different number of shots on the HellaSwag task using the KOBEST dataset. The evaluation metric used is the F1 score.
According to the table, our 12.8B model achieves the highest scores for the 5-shot, 10-shot, and 50-shot scenarios, with F1 scores of 0.6306, 0.6098, and 0.6118, respectively. However, in the 0-shot scenario, kakaobrain's kogpt model achieves the highest score of 0.5590. Overall, our 12.8B model demonstrates the best performance among the listed models on the HellaSwag task for the Korean language.
These results highlight the effectiveness of our 12.8B model in understanding and generating coherent responses within the context of the HellaSwag task. Its superior performance in the majority of the shot scenarios showcases its ability to handle complex language understanding tasks and generate contextually appropriate responses.

\subsection{BoolQ}

In the BoolQ task, which focuses on answering boolean questions, a thorough analysis of the results demonstrates that our models outperformed the others. Specifically, our largest model, Polyglot-ko-12.8B, achieved the highest F1 scores. This indicates that the model possesses exceptional accuracy in predicting answers to boolean questions. Conversely, SKT's ko-gpt-trinity model exhibited relatively lower F1 scores across all prompt numbers, while Facebook's XGLM model consistently underperformed. As a result, our models demonstrate strong performance in the BoolQ task.

\subsection{SentiNeg}

The SentiNeg task results are also presented in the left portion of Table \ref{tab:results_bq_sn}. This task focuses on sentiment analysis for negation detection. A comprehensive analysis of the results reveals that our models exhibited superior performance. In particular, our 12.8B model achieved the highest F1 scores, indicating its excellent ability to accurately detect negated sentiment. SKT's ko-gpt-trinity model also showed consistent improvement, while kakaobrain's KoGPT model demonstrated varied performance with slight increases or decreases in F1 scores depending on the prompt number. However, Facebook's XGLM model consistently displayed lower performance in the SentiNeg task.

Furthermore, during our investigation, we discovered that the default prompt used for this task introduced significant instability, particularly in zero-shot performance. To address this issue, we devised a modified prompt, which led to substantial improvements in the model's performance. For detailed results, please refer to the Appendix.
Overall, our models demonstrate strong performance in the SentiNeg task, showcasing their effectiveness in sentiment analysis and negation detection.

When analyzing the performance of Polyglot-Ko models only in the 5-shot evaluation across all four tasks, it becomes clear that the performance improves with increasing compute\cite{kaplan2020scaling} as depicted in Figure \ref{fig:5-shot}.

\section{Limitations and Disclaimers}
Polyglot-Ko has been primarily trained to optimize next token prediction, which makes it suitable for a wide range of tasks. However, it is crucial to acknowledge the potential for unexpected outcomes. While Polyglot-Ko strives to generate the most statistically likely response, it may not always provide the most accurate or factual answer. It is important to exercise caution when relying on the model's outputs.

Additionally, it is worth noting that Polyglot-Ko may generate content that is socially unacceptable or offensive. To mitigate this risk, we strongly recommend implementing a human curator or employing other filtering mechanisms to censor sensitive or inappropriate content.
Regarding the hardware used for training, it is important to mention that the models were trained on a hardware setup with relatively low TFLOPS compared to the upcoming versions of Polyglot that we are currently preparing. This resulted in longer training time and resources to complete the training process successfully.

Furthermore, we discovered mistakes in the data preprocessing phase during our experiments. Specifically, the data was incorrectly stripped of newlines, leading to a loss of document structure. This likely resulted in some information loss during the model training process. It is important to address this issue in future iterations to ensure the preservation of document structure and minimize information loss.
These considerations highlight the importance of continuously improving the training process and addressing any limitations or errors that arise. By doing so, we can enhance the performance and reliability of Polyglot-Ko for a wide range of tasks and applications.

\section{Conclusion}
Currently, we are actively working on training the new version of the Polyglot Korean language model. Our aim is to expand its capacity to eventually reach 40B parameters. This process has involved significant trial and error as we strive to enhance the performance and capabilities of the model.

Based on our experience and expertise in developing Korean language models, we have also embarked on the creation of two types of multilingual models. The first type is an East-Asian model, which includes Korean, Chinese, Japanese, Indonesian, Malay, Vietnamese, Thai, and English. This model aims to cater to the linguistic needs of countries in the East-Asian region.

The second type is a Romance model, which incorporates Spanish, Portuguese, French, Romanian, and Italian. This model is designed to support the linguistic requirements of Romance language-speaking countries.

By developing these multilingual models, we aim to democratize and promote access to language model technology across the globe. We believe that this will contribute to the advancement of research and academics in various countries, allowing users to leverage the power of language models for diverse applications and linguistic contexts. We are excited about the potential impact and benefits these models can bring to researchers, practitioners, and language enthusiasts worldwide.


\section*{Acknowledgements}
We are grateful to Stability AI for generously providing the essential computing resources that were instrumental in the successful execution of this project. Their support and infrastructure were crucial for training and evaluating our models.

Additionally, we would like to extend our appreciation to TUNiB for their invaluable contribution in providing a large-scale Korean dataset. This dataset played a pivotal role in the development and training of our language models, and we are grateful for their collaboration and partnership.

Finally, we would like to thank Stella Biderman for her valuable feedback on the paper, which greatly enhanced the quality and clarity of our work.


\bibliography{anthology,custom}
\bibliographystyle{acl_natbib}

\appendix

\section{Prompt Modification for SentiNeg Task}
\label{sec:appendix-sentineg}
In the case of SentiNeg, we observed that the prompt used in the KoBEST paper employed a simple classification task. However, the range of questions encompassed a wide spectrum and often exhibited ambiguity. Consequently, we made arbitrary modifications to the prompt in order to obtain results. The results of these prompt modifications are presented in the left portion of Table \ref{tab:msn_wic} and Figure \ref{fig:msentineg}. It is evident from the results that the scores achieved with the modified prompt are substantially higher on average than those achieved with the original prompt. This demonstrates the importance of prompt design and customization in achieving better performance in the SentiNeg task.

\section{Results for the WiC task}
\label{sec:appendix-wic}
\begin{figure*}
  \centering
  \subfloat{\includegraphics[width=.5\textwidth]{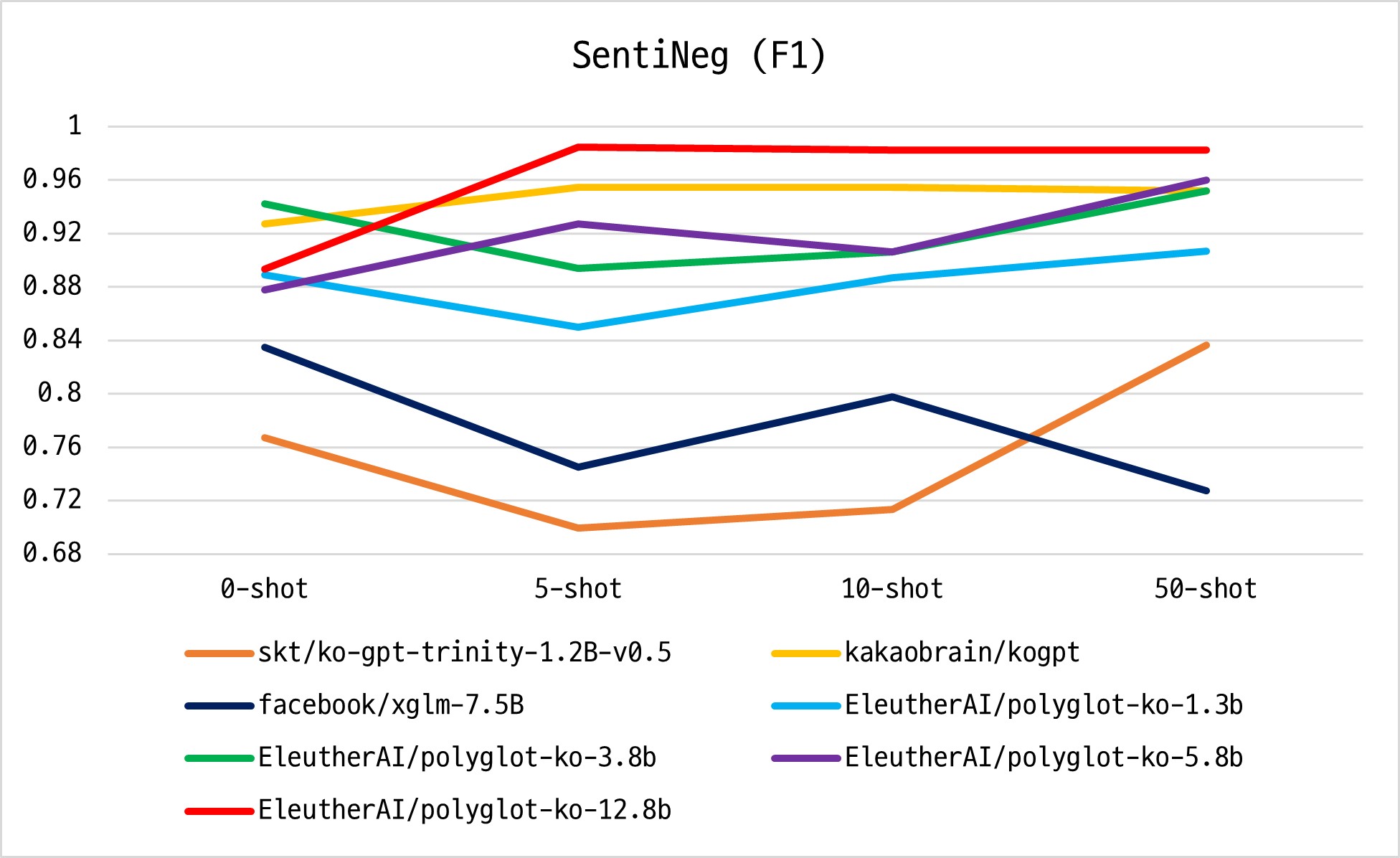}\label{fig:msentineg}}
  \hfill
  \subfloat{\includegraphics[width=.47\textwidth]{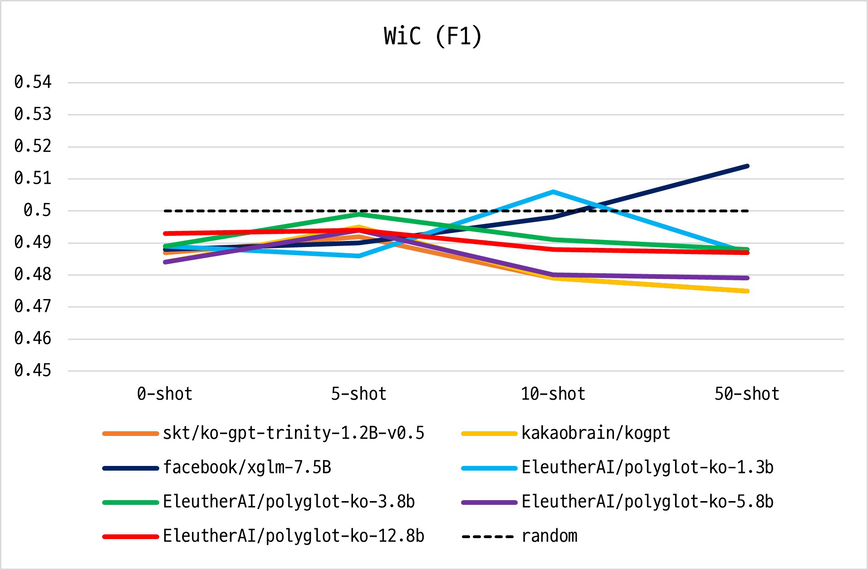}\label{fig:wic}}
  \caption{The performance metrics of the SentiNeg task with a modified prompt (left) and WiC task (right) using the KOBEST dataset, with the F1 scores for SentiNeg while the accuracy for WiC to show random performance.}
  \label{fig:wic_msent}
\end{figure*}

The results of the WiC task are presented in the right portion of Table \ref{tab:msn_wic} and visualized in Figure \ref{fig:wic}. Notably, all models demonstrated random performance in this task. We utilized the accuracy metric as it provided a more straightforward evaluation of performance, particularly for random performance. The consistent random performance across all models indicates that they encountered challenges in making accurate predictions for the WiC task. Therefore, further investigation and improvements are necessary to enhance the models' capability to effectively tackle this task.


\begin{table*}
    \centering
    \begin{tabular}{lccccccccc}
        \toprule
        \multicolumn{2}{c}{} & \multicolumn{4}{c}{\textbf{Modified SentiNeg} (n=shot)} & \multicolumn{4}{c}{\textbf{WiC}} \\
        \cmidrule(lr){3-6} \cmidrule(lr){7-10}
        \textbf{Model} & \textbf{Params} & n=0 & n=5 & n=10 & n=50 & n=0 & n=5 & n=10 & n=50 \\
        \midrule
            Ko-GPT-Trinity & 1.2B & 0.767 & 0.699 & 0.713 & 0.836 & 0.487 & 0.492 & 0.479 & 0.475 \\
            koGPT & 6B & 0.927 & 0.955 & 0.955 & 0.952 & 0.484 & 0.495 & 0.479 & 0.475 \\
            xGLM & 7.5B & 0.835 & 0.745 & 0.797 & 0.727 & 0.488 & 0.490 & 0.498 & \textbf{0.514} \\
            Polyglot-Ko (ours) & 1.3B & 0.889 & 0.850 & 0.887 & 0.907 & 0.489 & 0.486 & \textbf{0.506} & 0.487 \\
            Polyglot-Ko (ours) & 3.8B & \textbf{0.942} & 0.894 & 0.906 & 0.952 & 0.489 & \textbf{0.499} & 0.491 & 0.488 \\
            Polyglot-Ko (ours) & 5.8B & 0.878 & 0.927 & 0.906 & 0.960 & 0.484 & 0.494 & 0.480 & 0.479 \\
            Polyglot-Ko (ours) & 12.8B & 0.893 & \textbf{0.985} & \textbf{0.982} & \textbf{0.982} & \textbf{0.493} & 0.494 & 0.488 & 0.487 \\
        \bottomrule
    \end{tabular}
    \caption{Comparative results of the performance on the SentiNeg task with a modified prompt and the WiC task, with the F1 scores for SentiNeg while the accuracy for WiC to show random performance.}
    \label{tab:msn_wic}
\end{table*}

\end{document}